%% file: main.tex
\documentclass[aps,prl,twocolumn,nofootinbib]{revtex4-2}
\usepackage{graphicx}
\usepackage{mathrsfs}
\usepackage{color}
\graphicspath{{figures/}{fig/}}
\usepackage{amsmath}
\usepackage{amssymb}
\usepackage{bm}
\usepackage{slashed}
\usepackage{epsfig}
\usepackage{enumitem}
\usepackage{amsfonts}
\usepackage{epstopdf}
\usepackage{hyperref}
\usepackage{bbm}
\usepackage{textcomp}
\usepackage{color}
\usepackage{xcolor}
\usepackage[ruled,linesnumbered]{algorithm2e}

\newcommand{\sect}[1]{{\it \textbf{#1.} --- }}

\allowdisplaybreaks[3]

\setlist[description]{leftmargin=*,labelindent=\parindent}
\begin{document}

\title{AI-Newton: A Concept-Driven Physical Law Discovery System without Prior Physical Knowledge}

\author{You-Le Fang}
\email{eden@stu.pku.edu.cn}
\affiliation{School of Physics, Peking University, Beijing 100871, China}

\author{Dong-Shan Jian}
\email{dsjian@stu.pku.edu.cn}
\affiliation{School of Physics, Peking University, Beijing 100871, China}

\author{Xiang Li}
\email{lix-PHY@pku.edu.cn}
\affiliation{School of Physics, Peking University, Beijing 100871, China}

\author{Yan-Qing Ma}
\email{yqma@pku.edu.cn}
\affiliation{School of Physics, Peking University, Beijing 100871, China}
\affiliation{Center for High Energy Physics, Peking University, Beijing 100871, China}

\date{\today}

\begin{abstract}
While current AI-driven methods excel at deriving empirical models from individual experiments, a significant challenge remains in uncovering the common fundamental physics that underlie these models---a task at which human physicists are adept. To bridge this gap, we introduce AI-Newton, a novel framework for concept-driven scientific discovery. Our system autonomously derives general physical laws directly from raw, multi-experiment data, operating without supervision or prior physical knowledge. Its core innovations are twofold: (1) proposing interpretable physical concepts to construct laws, and (2) progressively generalizing these laws to broader domains. Applied to a large, noisy dataset of mechanics experiments, AI-Newton successfully rediscovers foundational and universal laws, such as Newton's second law, the conservation of energy, and the universal gravitation. This work represents a significant advance toward autonomous, human-like scientific discovery.
\end{abstract}

\maketitle

\sect{Introduction}
\label{sec:intro}
For centuries, the ultimate goal of fundamental physics research has been to describe a wide range of phenomena through a small number of discovered laws.
Advances in artificial intelligence (AI) have made AI-driven scientific discovery a highly promising new paradigm~\cite{xu2021artificial}. Although AI has achieved remarkable results in tackling domain-specific challenges~\cite{wang2023scientific,zhang2023artificial}, the ultimate aspiration from a paradigm-shifting perspective still lies in developing reliable  AI systems capable of autonomous scientific discovery directly from a large collection of raw data without supervision \cite{lu2024ai,reddy2024towards}. 

Current approaches to automated physics discovery focus on individual experiments, employing either neural network (NN)-based methods~\cite{schmidt2009distilling,brunton2016discovering,champion2019data,wu2019toward,greydanus2019hamiltonian,cranmer2020lagrangian,de2020discovery,liu2021machine,karniadakis2021physics,liu2022machine,camps2023discovering,cornelio2023combining,lemos2023rediscovering,liu2023discovering,cory2024evolving,zheng2018unsupervised,tegmark2019latent,iten2020discovering,chen2022automated,li2023metalearning} or symbolic techniques~\cite{udrescu2020ai,udrescu2020ai20,bendinelli2023controllable,tenachi2023deep,tian2024sym,cranmer2023interpretable,du2024large,romera2024mathematical}. By analyzing data from a single experiment, these methods can construct a specific model capable of predicting future data from the same experiment; if sufficiently simple, such a model may even be expressed in symbolic form~\cite{valipour2021symbolicgpt,chu2023neural,mevznar2023efficient}. Although these methods  represent a crucial and successful stage towards automated scientific discovery, they have not yet reached a discovery capacity comparable to that of human physicists. 

Human scientists advance further by discerning common patterns across specific models from different experiments and, on that basis, formulating general models that account for data from all such experiments. For instance, Newtonian mechanics provides a unifying and interpretable framework by defining meaningful physical concepts and formulating general laws that are valid across diverse phenomena. Therefore, a central challenge for the AI-driven physics discovery field is to evolve beyond problem-specific model fitting towards AI systems capable of discovering knowledge that is inherently generalizable and universally applicable.

In this Letter, we present AI-Newton, a concept-driven discovery system, which is designed for the critical question: how to extract concepts and general laws from problem-specific models. AI-Newton integrates an autonomous discovery workflow which is fundamentally built upon plausible reasoning and physical concepts. Given a collection of physical experiments, AI-Newton can gradually formulate a set of general laws applicable across a wide problem scope with neither supervision nor any prior physical knowledge. As a proof-of-concept implementation\footnote{Code available at \url{https://github.com/Science-Discovery/AI-Newton}}, by applying it to 46 different classical mechanics experiments, it can rediscover Newton's second law, energy conservation, law of gravitation and others in classical mechanics.

\begin{figure*}[t]
	\centering
 \includegraphics[width=0.8\textwidth]{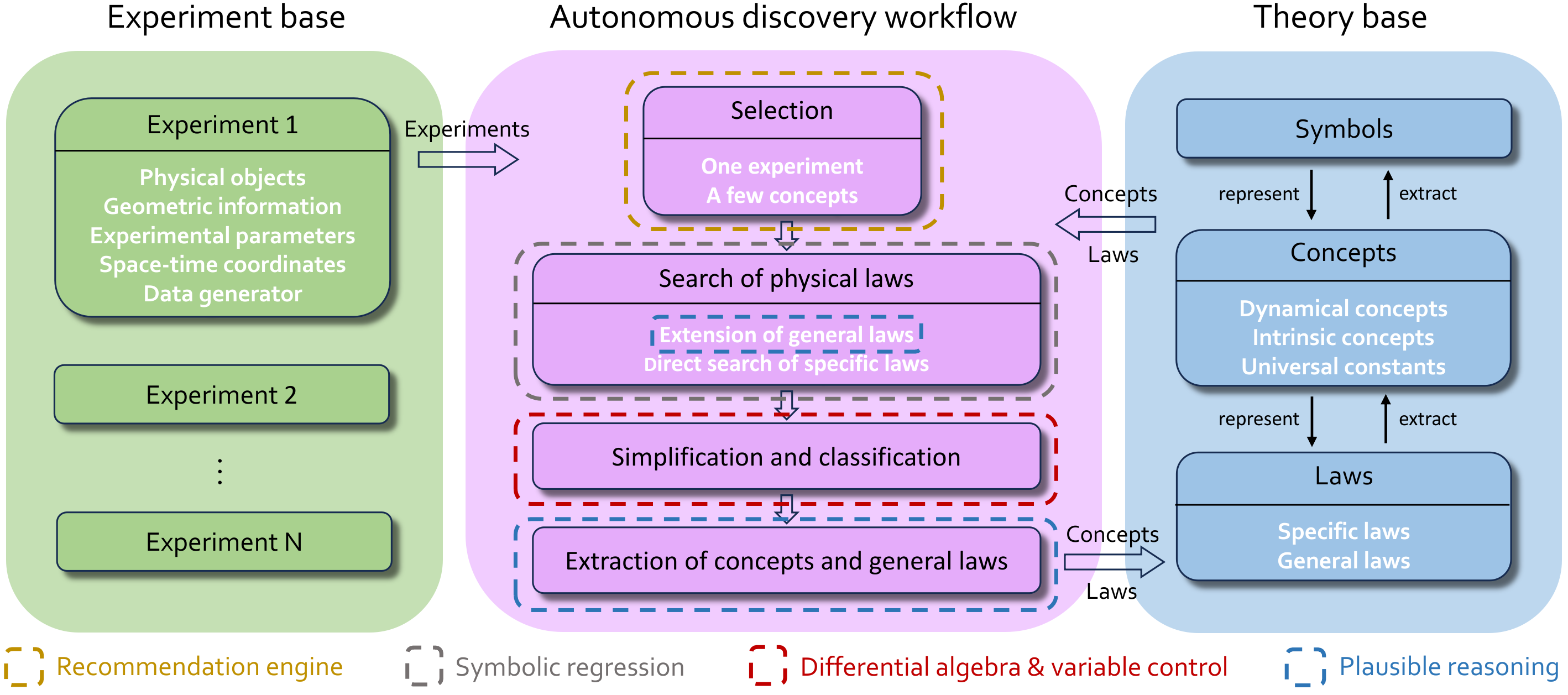} 
 \caption{AI-Newton's experiment base, theory base, and autonomous discovery workflow.}
	\label{fig:overview} 
\end{figure*}

\sect{Knowledge base and knowledge representation}
AI-Newton contains an experiment base and a theory base, as shown in Fig.~\ref{fig:overview}. 
The experiment base stores physical experiments and corresponding simulated data generators. The inputs for each experiment include only the physical objects involved, geometric information, experimental parameters, and space-time coordinates, which define an experiment. To emphasize that no prior physical knowledge is used, all other concepts, such as mass or energy, are autonomously discovered in AI-Newton. The output of each experiment is simulated data with statistical errors.
 
The theory base stores physical knowledge explicitly in an interconnected library of symbols, concepts, and laws. This design mirrors how human physicists construct concise, universal laws from conceptual building blocks. In contrast to prior work, which interprets latent features in NNs as physical concepts~\cite{wang2019emergent,iten2020discovering,li2024discover}, AI-Newton represents concepts and laws in an explicit, symbolic form. This greatly enhances interpretability and makes the acquired knowledge easier to transfer to new problems. Moreover, the introduction of powerful intermediate concepts allows complex physical laws to be expressed concisely, which in turn makes them more amenable to discovery through techniques like symbolic regression (SR)\cite{udrescu2020ai,udrescu2020ai20,bendinelli2023controllable,tenachi2023deep,tian2024sym,cranmer2023interpretable,du2024large,romera2024mathematical,valipour2021symbolicgpt,chu2023neural,mevznar2023efficient}. Initially, the concept layer contains only space-time coordinates; new concepts are autonomously defined and registered using a dedicated physical domain-specific language (DSL).
(See Supplemental Materials (SMs)\cite{supplement} for details.)

A robust knowledge representation is crucial because our goal is for the AI to discover generalizable knowledge across diverse systems, which requires transferring knowledge between different problems. To achieve this, we designed a physical DSL with a well-defined structure. This DSL not only formulates equations but also encodes the properties of physical objects and the relationships between physical quantities.
For instance, given the known concepts of coordinate $x$ and time $t$, the velocity of a ball can be defined in the DSL as:
\begin{equation}
    C_{1} := \forall i \text{: Ball}, \,\mathrm{d}x[i]/\mathrm{d}t,
\end{equation}
where $i$ indexes the balls and $C_1$ denotes the symbol of velocity, with the subscript $1$ varying across tests. In addition to dynamical concepts like velocity, the system also automatically identifies two other types: intrinsic concepts (e.g., mass, spring constant), which depend solely on specific physical objects, and universal constants (e.g., the gravitational constant), which are independent of all other quantities. Both are defined by documenting their measurement procedures. For example, mass of a ball could be defined as:
\begin{equation}
\begin{split}
    C_{2} := & \forall i \text{: Ball}, \text{Intrinsic}[\\
    &\text{ExpName}(o_1 \rightarrow i, o_2 \to s), \text{L}[s] - \text{L}_0[s]],
\end{split}
\end{equation}
where ExpName is the name of an experiment. In this experiment, the measured ball $i$ is suspended from a fixed spring $s$, and the spring elongation $\text{L}[s] - \text{L}_0[s]$ serves as the measurement of the mass. Recording the measurement procedures of intrinsic concepts is essential, since it allows the value of an intrinsic property to be retrieved by invoking its defining experiment, ensuring conceptual consistency across different problems.

These explicit concepts serve as the building blocks for the laws layer, which stores discovered physical laws, such as conserved quantities and dynamical equations. The laws are categorized into specific laws (valid for one experiment with specific forms) and general laws (valid across diverse experiments with general forms).  Within this framework, prior research in AI-driven physics discovery has concentrated on identifying specific laws. The introduction of general laws enables AI-Newton to simultaneously describe physics in various complex systems with compact and concise formulations. For instance, consider a system with a ball on an inclined plane connected to a fixed end via a spring. By applying the general law discovered by AI-Newton (Newton's second law in the $x$-direction):
\begin{equation}
    \forall i: \text{Ball},\,m_ia_{i,x}+ (\nabla_i V_g)_x + (\nabla_i V_k)_x  = 0,
\end{equation}
the more complex dynamical equation of the ball can be concretely derived as:
\begin{equation}
\begin{split}
    &ma_{x} - \frac{c_{x} c_{z}}{c_{x}^{2} + c_{y}^{2} + c_{z}^{2}}mg \\
    +& \frac{ \left[\left(c_{y}^{2} + c_{z}^{2}\right) x - c_{x} \left( c_{y} y + c_{z} z\right)\right]}{\left(c_{x}^{2} + c_{y}^{2} + c_{z}^{2}\right) L}k\Delta{L}=0,
\end{split}
\end{equation}
where $(c_x, c_y, c_z)$ is the normal vector defining the inclined plane. For multi-object systems,  concrete dynamical equations can be much more complex than the general laws, making them hard to be obtained using previous symbolic approaches. These cases highlight the efficacy of our concept-driven hierarchical approach.
 
\sect{Autonomous discovery workflow}
The autonomous discovery workflow in AI-Newton continuously distill knowledge---expressed as physical concepts and laws---from experimental data, as shown in Fig.~\ref{fig:overview}. Plausible reasoning, a method based on rational inference from partial evidence~\cite{polya1990mathematics1,polya1990mathematics2}, is the key to discovering knowledge. Unlike deductive logic, it produces contextually reasonable rather than universally certain conclusions, mirroring scientific practice where hypotheses precede rigorous verification.

The workflow initiates each trial by selecting an experiment and a few concepts from the theory base. This selection is governed by a recommendation engine that integrates a UCB-inspired value function~\cite{lai1985asymptotically,lai1987adaptive,agrawal1995sample,katehakis1995sequential,auer2002using} with a dynamically adapted NN. The NN's architecture is updated in real-time to favor  configurations that lead to efficient knowledge extraction. This mechanism enables the system to emulate human-like learning, naturally balancing the trade-off between exploration and exploitation.

To ensure the workflow establishes foundational knowledge before tackling complex experiments, we introduce an era-control strategy. Within a given era, every trial must conclude within a specific wall-clock time limit. If no new knowledge is acquired after a sufficient number of trials, the system advances to a new era with an exponentially increased time limit. Consequently, this strategy keeps the system focused on simpler experiments in the early phases. (See SMs\cite{supplement} for more details.)

The next step of each trial is to explore new laws from the selected experiment and concepts. Specific laws can be discovered through direct searching for relations among the selected concepts within the allowed operational space, which is nothing but SR.  Our SR implementation combines direct instantiation-verification and PCA-based differential polynomial regression\cite{Pearson01111901,wang2016discovering,kiwata2019deriving,yevick2021conservation}.
Furthermore, new general laws may emerge by extending existing ones through plausible reasoning. The core idea of plausible reasoning here is that, if a general law holds across multiple experiments but fails in the current one, there is a possibility to derive a valid modified law by adding simple terms to the original formulation via SR. For instance, while kinetic energy conservation governs elastic collisions, it fails in spring systems. Through plausible reasoning, AI-Newton introduces additional terms (elastic potential) to restore conservation. 
Mirroring human research practice, the system heuristically leverages existing general laws and selected concepts to search for physical laws that explain new experimental data.

The aforementioned process may generate redundant knowledge causing an explosion in both the theory base and search space that severely hinders continuous discovery under limited resources. To address this, AI-Newton simplifies physical laws into minimal representations in each trial. For the example shown in this paper, we employ the Rosenfeld Gr\"obner algorithm~\cite{boulier1995representation,boulier2009computing,maplesoft2024differentialalgebra,maple2024} from differential algebra to perform the simplification (See SMs\cite{supplement} for more details). Furthermore, through controlled-variable analysis, AI-Newton numerically identifies the dependencies of relations on physical objects and experimental parameters, using these dependencies as the basis for classification.

After identifying new laws, AI-Newton extracts new concepts from the processed results through plausible reasoning: a conserved quantity in the current experiment suggests broader utility, triggering its extraction as a new concept. Similarly, it proposes new general laws from directly-searched specific laws that also hold in multiple other experiments. All accumulated knowledge are updated to the theory base.

\begin{figure*}[t]
	\centering
	\includegraphics[width=0.8\textwidth]{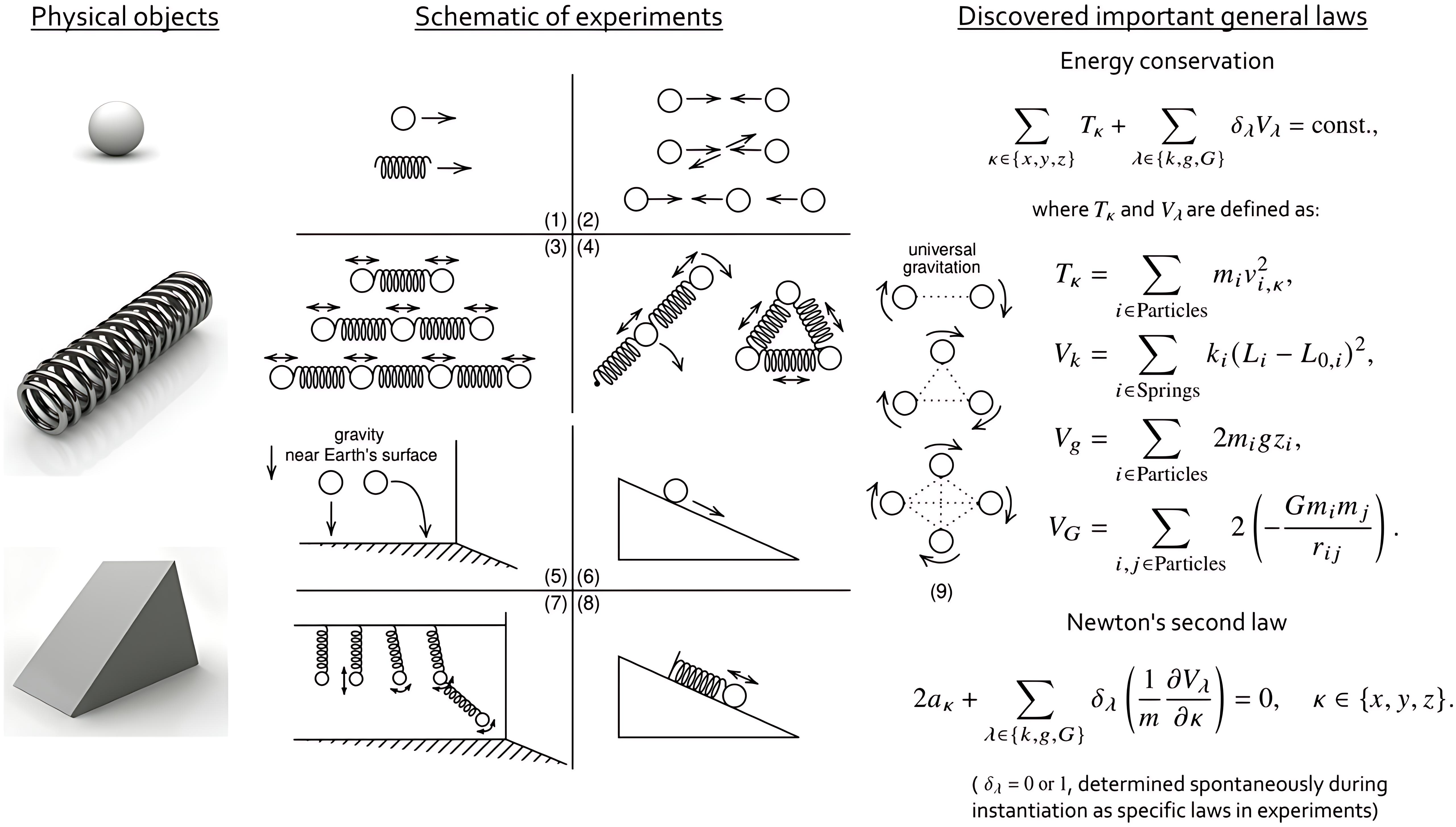} 

    \caption{Schematic of tested experiments and main general laws discovered. Some complex configurations are omitted for clarity. See text for details.}
	\label{fig:test cases}
\end{figure*}

\sect{Rediscovering Laws of Newtonian Mechanics}
To evaluate AI-Newton's performance, we apply it to Newtonian mechanics problems, focusing on a set of 46 predefined experiments. These experiments involve three primary types of physical objects: balls (either small balls or celestial bodies), springs, and inclined planes. The experiments are designed to investigate both isolated and coupled systems, as illustrated in Fig.~\ref{fig:test cases}, including:
\begin{enumerate}
    \item Free motion of individual balls and springs;
    \item Elastic collision of balls;
    \item Coupled systems demonstrating translational vibrations, rotational oscillations, and pendulum-like motions;
    \item Gravity-related problems, such as projectile motion and motion on inclined planes, along with complex spring-ball systems;
    \item Celestial mechanics problems involving gravitational interactions.
\end{enumerate}
The complexities of experiments are systematically increased by varying the number of physical objects and spatial dimensions, encompassing high-degree-of-freedom problems such as coupled oscillations of chained 2-ball-2-spring systems on inclined planes, rotational dynamics of 4-ball-4-spring systems, and other complex configurations. To simulate realistic experimental conditions, all test data are generated by solving differential equations and incorporating Gaussian-distributed errors. This comprehensive experimental setup covers three types of forces in Newtonian mechanics, elastic forces, gravity near Earth's surface, and universal gravitational forces, while incorporating realistic measurement uncertainties. In this way, it enables rigorous evaluation of AI-Newton's capability to discover physical laws from noisy experimental data. 

We evaluated the performance of our proof-of-concept implementation  on an Intel Xeon Platinum 8370C (128 threads @ 3.500GHz) platform with NVIDIA A40 GPU, configured with 64 cores for parallel processing. With max trials set to 1200 and an average runtime of 48 hours, the system demonstrated robust knowledge discovery capabilities, identifying approximately 90 physical concepts and 50 general laws on average across the test cases. The discoveries include significant general laws such as energy conservation and Newton's second law along with their relevant concepts, as shown in Fig.~\ref{fig:test cases}, providing complete explanatory for all experiments covering systems from simple to high-degree-of-freedom complex configurations. 

\begin{figure}[h]
	\centering
\includegraphics[width=0.45\textwidth]{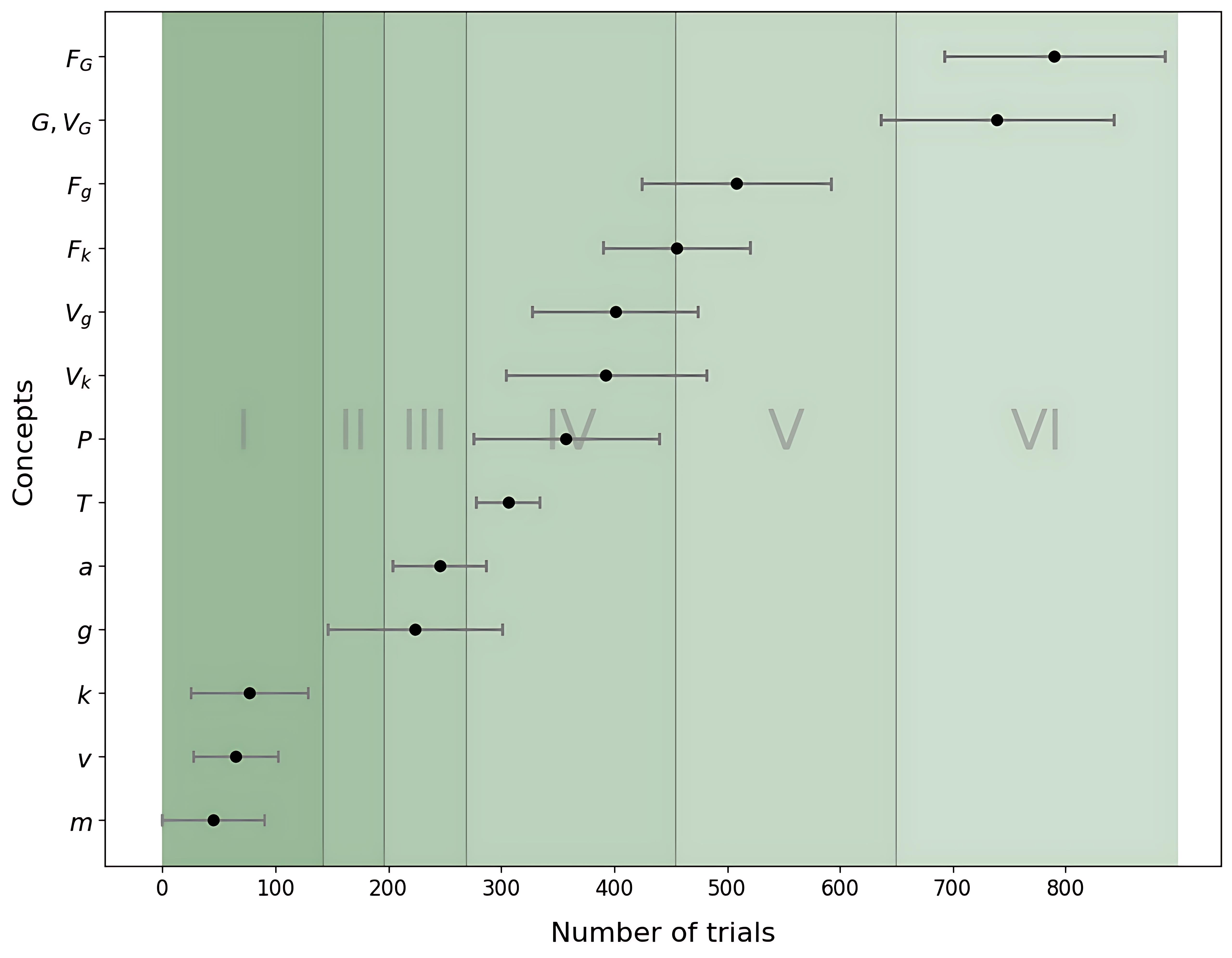}     \caption{Statistical analysis of concept discovery timing on 10 test cases, recording the mean and standard deviation of discovery timings for key concepts. Number of trials means the number of analysis trial attempt has been done, not distinguishing which experiment. Roman numerals (I, II, ...) in the background indicate the eras defined by the era-control strategy.}
	\label{fig:results} 
\end{figure}

Statistical discovery progression on 10 test cases is illustrated in Fig.~\ref{fig:results}, showing the timing distribution of important concept discoveries. This discovery progression exhibits an incremental pattern, where AI-Newton first explores simple concepts (e.g., mass) before advancing to more complex ones (e.g., force). 
For instance, gravitational acceleration $g$ is defined as a constant by analyzing free-fall or projectile motion, where the vertical acceleration $a_z$ of the ball is invariant. In experiments with elastic collisions between balls, conservation of kinetic energy $T$ is discovered and proposed as a general law. 
Through plausible reasoning, elastic potential energy $V_k$, gravitational potential energy near Earth's surface $V_g$, and universal gravitational potential energy $V_G$ are progressively defined when trying to apply the conservation of kinetic energy to inelastic experiments. These are then incorporated with kinetic energy conservation to ultimately formulate the complete law of energy conservation. The discovery of Newton's second law follows an analogous progression: it is first proposed in a simple experimental context and then generalized through plausible reasoning.

It is important to emphasize that the system is able to independently discover and unify fundamental concepts from disparate physical contexts. For instance, AI-Newton can derive the concept of `mass' through two distinct experimental routes: from the static elongation of a spring under gravity (defining gravitational mass, $m_g$) and from the experiment of a horizontal spring-mass oscillation system (defining inertial mass, $m_i$). Critically, the system then autonomously verify the numerical equivalence of $m_g$ and $m_i$, effectively indicating a cornerstone of general relativity---the weak equivalence principle---from raw data alone.

\sect{Summary} 
We introduce AI-Newton, a novel framework for the autonomous discovery of general physical laws from raw data across a large set of experiments, without supervision or pre-existing physical knowledge. This approach transcends current AI-driven methods, which are limited to extracting specific laws from individual experiments. Our main contributions are based on plausible reasoning, enabling us to: (1) propose physical concepts from the extracted laws; and (2) extend an existing general law by adding new terms, thereby adapting it to describe a wider range of experiments. Introducing interpretable physical concepts allows discovered laws to remain concise, making them more tractable for SR to identify. Furthermore, iteratively constructing general laws from existing ones enables a gradual, scalable discovery process. Applied to a large, noisy dataset of mechanics experiments, AI-Newton successfully rediscovers foundational laws, including Newton’s second law, the conservation of energy, and the law of universal gravitation.
This work thus offers a promising pathway toward building AI systems capable of contributing to frontier scientific research.

As a first step, we employ AI-Newton to rediscover known physical laws---a task where direct reliance on large language models (LLMs) is unsuitable, as they already possess this knowledge. In future applications to frontier science, however, the DSL, the recommendation engine and the plausible reasoning components of the framework could be replaced or augmented by LLMs. This integration would grant the system direct access to all existing knowledge, enabling a more informed and efficient discovery process.

\begin{acknowledgments}
We would like to thank Hong-Fei Zhang for early participant of the project and many valuable discussions. This work is supported by the National Natural Science Foundation of China (No. 12325503), and the High-performance Computing Platform of Peking University.
\end{acknowledgments}

\input{paper.bbl}


\end{document}

%% file: paper.bbl
\providecommand{\href}[2]{#2}\begingroup\raggedright\endgroup